\documentclass[11pt]{article}
\usepackage{newtxtext,newtxmath}

\usepackage[preprint]{acl}

\usepackage{times}
\usepackage{latexsym}

\usepackage[T1]{fontenc}

\usepackage[utf8]{inputenc}

\usepackage{inconsolata}

\usepackage{graphicx}

\usepackage{microtype}
\usepackage{hyperref}
\usepackage{url}
\usepackage{enumitem}
\usepackage{booktabs}
\usepackage{xcolor}

\usepackage{amsmath}
\usepackage{algorithm}
\usepackage{algpseudocode}
\usepackage{multirow}
\usepackage{multicol}
\usepackage{hyperref}

\usepackage{arydshln}
\usepackage{booktabs}
\usepackage{stfloats}

\usepackage{titlesec}
\usepackage{booktabs}

\usepackage{adjustbox}
\usepackage{changepage} 
\usepackage{lipsum} 

\usepackage{wrapfig}
\usepackage{lipsum}
\usepackage{listings}
\usepackage{tcolorbox}

\usepackage{amssymb} 
\usepackage{amsmath}
\usepackage{cuted}
\usepackage{amsfonts}
\usepackage{verbatim}

\tcbuselibrary{listings,skins} 
\usepackage{caption}
\usepackage{subcaption}
\usepackage{color}
\definecolor{lightred}{rgb}{1, 0.6, 0.6}  

\tcbset{
    casebox/.style={
        colback=white,
        colframe=blue!40,
        coltitle=white,
        fonttitle=\small\bfseries,
        fontupper=\normalsize,
        title=#1,
        boxrule=1pt,
        arc=4pt,
        top=5pt,
        bottom=5pt,
        left=10pt,
        right=10pt,
        width=\linewidth,
    },
    smallcasebox/.style={
        casebox, 
        title=#1,
        fontupper=\small,     
    },
     tinycasebox/.style={
        casebox, 
        title=#1,
        fontupper=\tiny,     
    }
}

\newtcbox{\rndlabel}{
    on line,
    colback=blue!10, 
    colframe=blue!30, 
    coltext=black, 
    boxrule=0.5pt, 
    arc=3pt, 
    left=3pt,
    right=3pt,
    top=1pt,
    bottom=1pt,
    fontupper=\small\bfseries, 
}

\newtcbox{\tinyrndlabel}{
    on line,
    colback=blue!10, 
    colframe=blue!30, 
    coltext=black, 
    boxrule=0.5pt, 
    arc=3pt, 
    left=3pt,
    right=3pt,
    top=1pt,
    bottom=1pt,
    fontupper=\tiny\bfseries, 
}

\newtcbox{\rndlabelalt}{
    on line,
    colback=yellow!20, 
    colframe=yellow!30, 
    coltext=lightred, 
    boxrule=0.5pt, 
    arc=3pt, 
    left=3pt,
    right=3pt,
    top=1pt,
    bottom=1pt,
    fontupper=\small\bfseries, 
}

\usepackage{listings}
\definecolor{myred}{RGB}{178, 34, 34} 
\definecolor{mygreen}{RGB}{34,139,34}   
\definecolor{myred2}{RGB}{237, 211, 210} 
\definecolor{mygreen2}{RGB}{198, 232, 206} 
\definecolor{myblue2}{RGB}{218,232,252}
\definecolor{codegreen}{rgb}{0,0.6,0}
\definecolor{codegray}{rgb}{0.5,0.5,0.5}
\definecolor{codepink}{RGB}{252, 142, 172}
\definecolor{codepurple}{rgb}{0.58,0,0.82}
\definecolor{backcolour}{RGB}{245,245,245}

\lstdefinestyle{mystyle}{
    backgroundcolor=\color{backcolour},   
    commentstyle=\color{magenta},
    keywordstyle=\color{blue},
    numberstyle=\tiny\color{codegray},
    stringstyle=\color{codepurple},
    basicstyle=\fontfamily{\ttdefault}\footnotesize,
    breakatwhitespace=false,         
    breaklines=true,                 
    keepspaces=true,    
    frame=single,
    numbersep=5pt,                  
    showspaces=false,                
    showstringspaces=false,
    showtabs=false,                  
    tabsize=2,
    classoffset=1, 
    keywordstyle=\color{violet},
    classoffset=0,
}
\lstdefinelanguage{JavaScript}{
  keywords={typeof, new, true, false, catch, function, return, null, catch, switch, var, if, in, while, do, else, case, break},
  keywordstyle=\color{blue}\bfseries,
  ndkeywords={class, export, boolean, throw, implements, import, this},
  ndkeywordstyle=\color{darkgray}\bfseries,
  identifierstyle=\color{black},
  sensitive=false,
  comment=[l]{//},
  morecomment=[s]{/*}{*/},
  commentstyle=\color{purple}\ttfamily,
  stringstyle=\color{red}\ttfamily,
  morestring=[b]',
  morestring=[b]"
}

\lstdefinestyle{sqlstyle}{
    language=SQL,
    basicstyle=\small\ttfamily,
    keywordstyle=\color{blue}\bfseries,
    commentstyle=\color{green!60},
    numbers=left,
    numberstyle=\tiny\color{gray},
    frame=single,
    framesep=5pt,
    rulecolor=\color{gray!30},
    breaklines=true,
    showstringspaces=false,
    xleftmargin=15pt,
    xrightmargin=5pt,
    backgroundcolor=\color{gray!5}
}

\lstdefinestyle{tinysqlstyle}{
    language=SQL,
    basicstyle=\tiny\ttfamily,
    keywordstyle=\color{blue}\bfseries,
    commentstyle=\color{green!60},
    numbers=left,
    numberstyle=\tiny\color{gray},
    frame=single,
    framesep=5pt,
    rulecolor=\color{gray!30},
    breaklines=true,
    showstringspaces=false,
    xleftmargin=5pt,
    xrightmargin=2pt,
    backgroundcolor=\color{gray!5}
}

\newtcolorbox{sqlhighlightbox}[2][]{
    listing only,
    listing options={style=sqlstyle, #1},
    colback=gray!5,
    colframe=gray!30,
    left=15pt,
    right=5pt,
    top=3pt,
    bottom=3pt,
    boxrule=0.5pt,
    title=#2,
}
 
\newcommand{\modelname}{RuCo-C}

\title{Beyond Query-Level Comparison: Fine-Grained Reinforcement Learning for Text-to-SQL with Automated Interpretable Critiques}

\author{Guifeng Wang, Yuanfeng Song, Meng Yang, Tao Zhu, \\ {\bf Xiaoming Yin, Xing Chen} \\
    ByteDance, Beijing, China \\
    \texttt{wgf1109@mail.ustc.edu.cn} \\
    \texttt{\{songyuanfeng, yangmeng.89\}@bytedance.com} \\
    \texttt{17310903278@163.com} \\
    \texttt{\{yinxiaoming, chenxng.xc\}@bytedance.com}
  }

\begin{document}
\maketitle
\begin{abstract}
Text-to-SQL, a pivotal natural language processing (NLP) task that converts textual queries into executable SQL, has seen substantial progress in recent years. However, existing evaluation and reward mechanisms used to train and assess the text-to-SQL models remain a critical bottleneck. 
Current approaches heavily rely on manually annotated gold SQL queries, which are costly to produce and impractical for large-scale evaluation. More importantly, most reinforcement learning (RL) methods in text-to-SQL leverage only the final binary execution outcome as the reward signal, a coarse-grained supervision that overlooks detailed structural and semantic errors from the perspective of rubrics. 
To address these challenges, we propose \textbf{\modelname}, a novel generative judge model for fine-grained, query-specific automatic evaluation using \textit{interpretable critiques} without human intervention. Our framework first automatically generates query-specific evaluation rubrics for human-free annotation, linking them to interpretable critiques. Subsequently, it integrates densified reward feedback through a ``progressive exploration'' strategy during the RL training process, which dynamically adjusts the rewards to enhance the model’s performance. Comprehensive experiments demonstrate that \modelname~outperforms existing methods in text-to-SQL evaluation, yielding significant performance gains. 
\end{abstract}

\section{Introduction}

As a pivotal natural language processing (NLP) task, text-to-SQL turns textual queries into executable SQL, has achieved significant advancement over the past few years~\cite{11095853,zhang2024natural}. Current text-to-SQL solutions focus on optimizing various components of the text-to-SQL workflows, including the pre-processing module (e.g., schema linking~\cite{guo2019towards, talaei2024chess}), the translation module~\cite{wang2020rat, ICLR2025_974ff7b5, li2025omnisqlsynthesizinghighqualitytexttosql} and the post-processing (e.g., output consistency~\cite{gao2024text, DBLP:journals/corr/abs-2403-09732}). More recently, various reinforcement learning (RL) methods have been proposed to enhance the reasoning and generalization capabilities of text-to-SQL. In terms of reward function design, some RL approaches primarily use execution accuracy (EX) as the feedback signal (e.g., SQL-R1~\cite{ma2025sqlr1trainingnaturallanguage}), while other works combines multiple reward functions such as n-gram similarity, syntax correctness, and executed table cell results (e.g., Reasoning-SQL~\cite{pourreza2025reasoning}, Think2SQL~\cite{papicchio2025think2sqlreinforcellmreasoning}).

Despite the aforementioned advancements, existing evaluation and reward mechanisms for training and assessing the text-to-SQL models remain a critical bottleneck. Specifically, current evaluation methods primarily utilize query-based pairwise comparisons between the predicted SQL query and the gold standard query (e.g., exact match~\cite{song2024revisiting}), execution matching~\cite{zhong2020semantic}, and embedding matching~\cite{zhan2025towards}). These approaches rely on manually annotated gold SQL queries, which are costly to produce and impractical for large-scale evaluation. 
Moreover, most RL approaches in text-to-SQL leverage only the final binary execution outcome as the reward signal \cite{pourreza2025reasoning,weng2025graph}, a coarse-grained supervision that ignores detailed structural and semantic errors in the intermediate reasoning steps. 

\begin{figure}[t!]
        \begin{tcolorbox}[smallcasebox={Critique Response}]
        \vspace{-5pt}
            \textbf{<think>}\\
        1. Did I correctly specify the column to filter the bond type? \\
        - Yes, the ``WHERE bond\_type = `triple''' clause correctly filters for triple bond types. \\
        2. Did I join the necessary tables to retrieve the required information?\\
        - \textcolor{red}{No, the query does not join the ``molecule'' table}, which is necessary to determine if the molecules are carcinogenic or not. The ``bond'' table has a ``molecule\_id'' column can be used to join with the ``molecule'' table.
        \\
        ...
            \\
            \textbf{</think>}
            \\[2pt]
            \textbf{<result>} \textcolor{red}{False} \textbf{</result>}
            \\[2pt]
            \textbf{<correctedSQL>}
            \begin{lstlisting}[style=sqlstyle, label=case:gt_sql]
SELECT b.bond_id, m.label 
FROM bond b 
JOIN molecule m ON b.molecule_id = m.molecule_id
WHERE b.bond_type = 'triple' \end{lstlisting}
            \textbf{</correctedSQL>}
        \end{tcolorbox}
    \vspace{-5pt}
    \caption{An example of the critique response illustrates the designed output format, which will be used for reward score in the RL training of the critique model.}
    \vspace{-10pt}
    \label{fig:critique_response}
\end{figure}

To address these challenges, we propose \textbf{\modelname} (\textbf{Ru}brics and reward \textbf{Co}nsistency \textbf{C}ritique model), a novel generative judge model that enables fine-grained, query-specific automatic evaluation using interpretable critiques without human intervention. 
First, we design \textit{interpretable critiques} by a self-asking approach and presented in Question-Answer (QA) pairs (i.e., the \textit{<think>} components in Figure~\ref{fig:critique_response}): the question identifies whether the items that need inspection for a user query are correct, while the answer elaborates on inspection conclusions based on specific evidence. This complete and detailed QA format fully embodies the evaluation rubrics. Then, we employ Supervised Fine-Tuning (SFT) to internalize rubric-based evaluation logic, aligning the model’s scoring behavior more closely with human expert judgments. Finally, in the RL training phase, we design fine-grained process rewards and a reward consistency mechanism.
Specifically, the Outcome Reward Model (ORM) is calculated based on execution accuracy, whereas the Process Reward Model (PRM) is primarily quantified based on our designed interpretable critique responses.
To account for task difficulty variations while preserving a consistent process reward, we dynamically tune the PRM's weight coefficient according to difficulty levels and correction status. These components densify the evaluation scores, enrich the feedback signals, and ultimately increase the stability of the training, all while improving the accuracy and robustness of the model's evaluation. Comprehensive experiments demonstrate that our \modelname~outperforms existing methods in text-to-SQL evaluation, significantly improving overall performance.

%

In a nutshell, our contributions can be summarized in three aspects:
\begin{itemize}
    \item \textbf{Evaluation Rubrics with Interpretable Critiques}: We automatically generate query-specific rubrics for human-free annotation, linking them to interpretable critiques.
    \item \textbf{Enhanced RL Training}: We integrate a novel rubric-related generative task into the RL training process, using a ``progressive exploration'' strategy to dynamically adjust rewards and enhance the model's inferential ability.
    \item \textbf{Novel Reward Design}: We construct the final reward function by combining ORM with PRM. In exploring PRM, we determine the combination weight of PRM by integrating the total number of rubric items and the specific responses guided by each rubric.
\end{itemize}

\section{Related Work}
\subsection{Text-to-SQL Evaluation Methods}

Existing evaluation methods for text-to-SQL mainly focus on query-based comparison evaluation~\cite{renggli2025fundamentalchallengesevaluatingtext2sql}, which can be systematically classified into three main categories: semantic matching~\cite{song2024revisiting}, execution matching~\cite{zhong2020semantic}, and embedding matching~\cite{zhan2025towards}. 
The semantic matching parses both gold and predicted SQL queries into an intermediate format (e.g., abstract syntax tree, AST~\cite{cao2023heterogeneous}) and  tests their equivalence via schema-based algebraic transformations~\cite{song2024revisiting}.
The execution matching assesses the equivalence by running two SQL queries on the same database instance and comparing the results. For instance, ~\newcite{zhong2020semantic} proposed Test-Suite Accuracy, which creates a small test suite of database via random generation to maximize the coverage of the gold query and calculates the denotation accuracy in this distilled set to estimate the semantic accuracy.
The embedding-based matching approaches, such as Codebertscore~\cite{zhou2023codebertscore} and FuncEvalGMN~\cite{zhan2025towards}, assess the semantic equivalence of SQL statements by computing the similarity between their vector representations. 


While existing query-level pairwise comparison methods have advanced the field of text-to-SQL assessment, they face several challenges: reliance on a single gold query, risk of spurious evaluations, and difficulty in gauging complex SQL semantics. By contrast, our designed generative judge model requires no gold SQL, which is better suited for online real-time scenarios. Furthermore, the generated critique information is interpretable and can be converted into more granular reward signals to guide the training of both the judge model itself and text-to-SQL models, thereby advancing innovation and precision in the field.


\begin{figure*}[th!]
    \centering
    \includegraphics[width=1.0\linewidth]{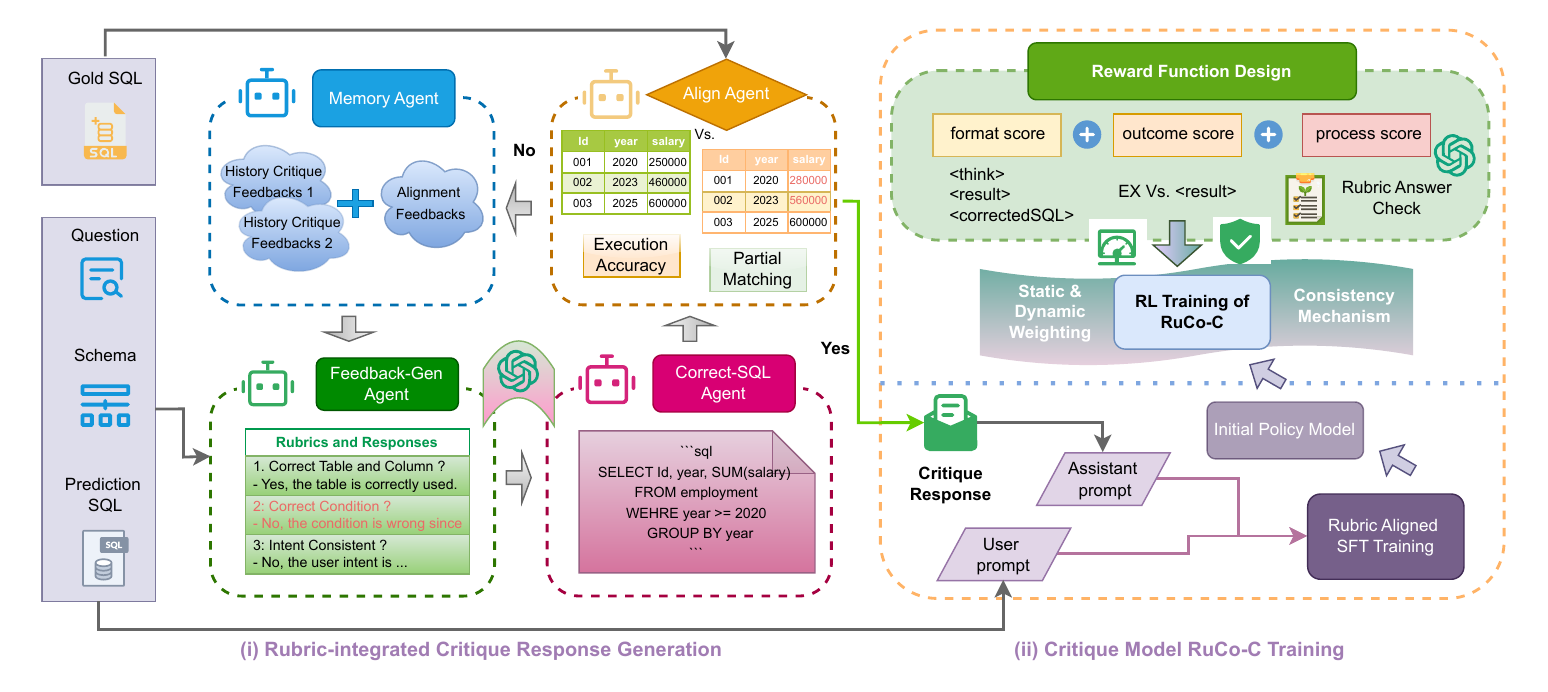}
    \caption{The working pipeline of our solution. It first automatically generates rubric-integrated critique response via a multi-agent framework. Then, it integrates densified reward feedbacks through a consistency mechanism combined with weighting strategy during RL training, dynamically adjusting rewards to enhance model performance.}
    \vspace{-0pt}
    \label{fig:solution_framework}
\end{figure*}

\subsection{Reward Models for RL Training}
Currently, the prevalent approach in most Reinforcement Learning with Verifiable Rewards (RLVR)~\cite{lambert2025tulu3pushingfrontiers, yue2025does} models is to combine rule scores with final outcome correctness scores as the reward score. For example, SQL-R1~\cite{ma2025sqlr1trainingnaturallanguage} considers syntactic correctness and result correctness through execution accuracy (EX) as the reward score in the reinforcement learning of the text-to-SQL task. Furthermore, Reasoning-SQL~\cite{pourreza2025reasoning} adopts a suite of reward functions (i.e., format, syntax check, schema linking, n-gram similarity, EX, and LLM-as-a-judge rewards) to produce a composite reward. Although Reasoning-SQL employed RL from AI Feedback, the calculation still relied on gold SQL queries. In contrast, this study focuses on training a point-wise LLM judge model, which can address the applicability in scenarios where gold SQL is unavailable.
Additionally, some general reward methods, such as DeekSeek-GRM~\cite{DBLP:journals/corr/abs-2504-02495}, JudgeLRM~\cite{chen2025judgelrmlargereasoningmodels}, and RM-R1~\cite{chen2025rmr1rewardmodelingreasoning}, employ reasoning models as generative reward models. 
However, the final rewards utilized by these methods remain relatively coarse-grained, as they mainly focus on the ultimate answer outcomes while neglecting the more informative reasoning processes. Consequently, they fail to meet the requirements for fine-grained evaluation of assessment tasks.

\section{Our Approach}

\subsection{Overview}
The framework of our proposed method \modelname~is illustrated in Figure~\ref{fig:solution_framework}, comprising two core components. First, we design a rubric-integrated critique response and adopt a multi-agent framework that synergizes generation and verification processes for data construction, ensuring data quality to support the subsequent rubric aligned SFT training of the critique model. Second, we treat the trained SFT model as the initial policy model for RL training, integrating fine-grained reward functions with a consistency mechanism that incorporates static and dynamic weighting. This yields a generative evaluation model capable of assessments without reliance on gold queries.


\subsection{Rubric-integrated Critique Response Generation}
For an interpretable evaluation model, we expect its output to include not only direct binary classification results but also specific evaluation criteria and corresponding rationale. Furthermore, for SQL statements identified as erroneous, we anticipate that the model can correct such errors. Therefore, we design an output structure comprising three key elements, referred to as a \textbf{critique response}. For more intuitive understanding, a concrete example of a critique response is provided in Figure~\ref{fig:critique_response}. The stepwise rubric-aligned evaluation reasoning process is presented between tags `<think>' and `</think>', a `False' judgment result indicates that the predicted SQL contains certain inconsistencies with the rubric's logic or grammar. 
The corrected SQL queries for erroneous predictions are enclosed within the `<correctedSQL>' and `</correctedSQL>'.

Unlike text-to-SQL generation models, our generative critique model requires not only user questions, corresponding database metadata, and model-predicted SQL, but also critique feedback (i.e. critique response). To reduce the cost of manual annotation, we propose a generative and verification based multi-agent architecture~\cite{sengupta2025magvmultiagentframeworksynthetic,tang2024synthesizing} to synthesize high-fidelity training datasets. 

Specifically, we implement an agent-based workflow in the data construction pipeline. 
As shown in Figure~\ref{fig:solution_framework}, the Memory Agent maintains a historical feedback buffer, which serves as a knowledge repository to inform subsequent iterative refinement processes. The Feedback-Gen Agent utilizes a self-questioning, stepwise reasoning paradigm to construct sample-specific critique metrics and associated reasoning trajectories. Meanwhile, the Correct-SQL Agent conducts systematic rectification of faulty SQL statements, ensuring data integrity and semantic consistency across the dataset. These agents operate in a synergistic manner, leveraging their complementary capabilities to optimize the end-to-end data generation pipeline.
Upon data generation, the Align Agent employs a two-stage validation mechanism, comprising the execution accuracy method~\cite{zhong2020semantic} and the partial matching algorithm~\cite{zhan2025towards} to filter instances where binary classification results align with ground truth labels or corrected SQL statements match the gold SQL.  

\subsection{Training of \modelname~Model}
The training of the \modelname~model is divided into two phases: Supervised Fine-Tuning (SFT) and Reinforcement Learning (RL).

\subsubsection{Rubric Aligned SFT Training}

\noindent \textbf{Problem Formulation.}
To train a generative critique model, we formulate the input and output of the model, which refer to the information in the user prompt and the LLM response. The input of sample $i$ contains the question, database schema and the predicted SQL, which is denoted as $X^{i} = \{{q, m, \hat{c}}\}$, $q$ represents the question, $m$ is corresponding database schema, $\hat{c}$ denotes the predicted SQL which is generated by the model being evaluated. The output is the critique response, which contains the stepwise rubric reasoning process $s^{i} = \{{s_1, s_2, ..., s_{N_{i}} \}}$, the classification result of predicted SQL $\hat{y}^{i}$, and the refined SQL statements $\tilde{c}^{i}$. Each reasoning information contains the specific rubric question $b$ and answers $a$, then the $k$-th reasoning step can be denoted as $s_k = \{{b_k, a_k}\}$. Since the rubric information for each sample is implicit in the stepwise reasoning process, a rubric-aligned critique model can be trained on the aforementioned input-output data.

\noindent \textbf{Supervised Fine-Tuning.}
In this study, we conduct SFT model to enhance the model’s capacity for instruction adherence and generation within the text-to-SQL judgment domain. We investigate two distinct SFT training strategies. The first one employs instructions focused solely on binary classification to determine the correctness of predicted SQL. The second one adopts rubric reasoning generation instructions, which prompt the development of rubric-based reasoning processes prior to reaching the final conclusion. Additionally, by appropriately controlling the ratio of positive to negative samples in the training data, we aim to obtain a critique model with unbiased category judgment.

\subsubsection{RL Training with Novel Reward Design}

\noindent \textbf{Reinforcement Learning.}
During the reinforcement learning stage, we integrate the Group Relative Policy Optimization (GRPO) algorithm~\cite{shao2024deepseekmathpushinglimitsmathematical} into our training framework. This algorithm eliminates reliance on a value model, boasts lower memory consumption, and allows for precise formulation of reward objectives. These qualities make it an ideal selection for efficiently optimizing the policy model of text-to-SQL evaluators.

For each natural language question paired with its corresponding database schema and the predicted SQL statements, the policy model generates a set of $G$  candidate critique responses $O = \{{o_1, o_2, ..., o_G\}}$ through the old policy $\pi_{old}$. These candidates are rigorously evaluated using a composite reward function that assigns specific scores. By focusing on the relative performance of these critique response candidates within the group, GRPO effectively computes rewards for each output, thereby directing policy updates to align with our predefined objectives:
\vspace{-10pt}
\begin{equation}
    \label{eq:opt_objective}
    \begin{aligned}
       & \mathcal{J}_{GRPO}(\theta) = \mathbb{E}_{\mathbf{u}\sim P(\mathbf{U}), \{o_i\}_{i=1}^{G}\sim \pi_{old}(O|\mathbf{u})} \\
    & \bigg[ \frac{1}{G} \sum_{i=1}^{G} \big(\min(r_i^{ratio} A_i, \text{clip}(r_i^{ratio}, 1-\epsilon, 1+\epsilon)A_i) \\
    & - \beta D_{KL}(\pi_{\theta} || \pi_{ref})\big) \bigg],
    \end{aligned}
\end{equation}
where $r_i^{ratio}=\frac{\pi_{\theta}(o_i|\mathbf{U})}{\pi_{old}(o_i|\mathbf{U})}$ denotes the importance sampling ratio, which measures the relative probability of generating critique output $o_i$ under the new policy $\pi_{\theta}$ as opposed to $\pi_{old}$. $A_i$ refers to the advantage computed exclusively from the relative rewards of outputs within each group. $\epsilon$ is clipping-related hyperparameter for stabilizing training, the hyperparameter $\beta$ controls the regularization of divergence between the trained policy $\pi_{\theta}$ and the reference policy $\pi_{ref}$.

\noindent \textbf{Reward Function Design.}
Conventional reward mechanisms often struggle to capture the intricacies of reasoning processes and are vulnerable to the influence of noisy data. To overcome these limitations and enhance model performance, we developed a multi-component reward design. This design aims to address key challenges, including ensuring model performance with limited outcome-based rewards, preventing oversight of intermediate reasoning steps, and mitigating the impact of noisy dataset labels. By dynamically combining outcome-based and process-aware signals, adjusted according to their credibility, we seek to explore whether a more comprehensive reward structure can better achieve improved performance.

Intuitively, the basic reward components centered on output format correctness $R_{format}$ and final binary judgment outcome $R_{out}$ are essential for meeting basic task requirements. However, recognizing the importance of intermediate rubric reasoning steps for robust model behavior, we incorporated a process reward (PR) by evaluating the correctness of answers in the critique model’s reasoning trajectory (aligned with the query and given information). This process reward refines the reward signal based on step-by-step reasoning accuracy, testing whether granular process supervision, conditional on a correct final judgment, can enhance model reliability. Specifically, the rubric-guided process reward score is designed as:
\begin{equation}
    R_{rubric} = 1 - \frac{1}{N} \sum_{i=1}^{N} \mathbb{I}(a_i \text{ is  incorrect}), 
\end{equation}
where the symbol $a_i$ means the answer in the $i$-th reasoning process. The final process reward score is derived by verifying the correctness of the answer at each individual step via strong LLMs.


Considering the consistency of final results and the partial guiding significance of reasoning processes, we meticulously design the coefficients for PR. The design is divided into two main components: static PR coefficient and dynamic PR coefficient, denoted as $\gamma_{s}$ and $\gamma_{d}$ respectively.

\begin{table*}[th!]
\centering
\setlength{\tabcolsep}{6pt}
\renewcommand{\arraystretch}{1.0}
    \resizebox{\textwidth}{!}{
    \begin{tabular}{llccc c ccc c ccc}
        \toprule
        \multirow{2}{*}{\textbf{Method Type}} & \multirow{2}{*}{\textbf{Method}} & \multicolumn{3}{c}{\textbf{Spider}} & & \multicolumn{3}{c}{\textbf{BIRD}} & & \multicolumn{3}{c}{\textbf{Spider-DK}}\\
         \cmidrule(lr){3-5} \cmidrule(lr){7-9} \cmidrule(lr){11-13}
        & & \textbf{AUC} & \textbf{ACC} & \textbf{F1} & &\textbf{AUC} & \textbf{ACC} & \textbf{F1} & & \textbf{AUC} & \textbf{ACC} & \textbf{F1} \\
        \midrule
        \multirow{2}{*}{Proprietary} & o3-mini & 67.68	& 66.24 & \textbf{72.32} & &  64.44 & 49.08 & 46.04 & & 75.54 & 67.61 & 60.52 \\
        & GPT-4.1 & \textit{\underline{68.80}} & 67.88 & 71.46 & & 67.55 &	58.08 &	48.64 & & \textit{\underline{76.28}} & 72.19 & 62.12 \\
        \midrule
        \multirow{6}{*}{Open-Source} & DeepSeek-R1-250120
 & \textbf{69.72} &	\textbf{69.40} & 69.93 & & 67.98	& 63.29	 & 49.33 & & \textbf{77.20} & \textit{\underline{77.41}} & \textbf{64.55}\\
        & Deepseek-V3-241226 &  64.67 & 62.90 & 71.04 & & 61.31	& 42.32 &	43.87 & & 72.85 &	63.29 & 57.70 \\
        & Kimi-K2 & 66.55 & 64.96 & \textit{\underline{71.88}} & & 63.25 & 47.70 & 45.12 & & 75.79 & 67.50 & 60.70 \\
        
        \cmidrule{2-13}
        & DeepSeek-Coder-6.7B-Instruct &  59.60 & 59.00 & 61.79 & & 55.42 &	48.94 & 38.11 & & 58.75 & 53.54 & 44.67\\
        & Qwen2.5-Coder-7B-Instruct & 56.05 & 53.71 & 66.64 & &
        54.25 &	33.19 & 39.49 & & 60.85 & 44.17 & 48.17\\
        & Qwen2.5-Coder-14B-Instruct & 59.68 & 57.54 & 68.50  & & 56.12 & 33.83 & 40.77 & & 66.04 &	52.96 &	51.77 \\
        \midrule
        \multirow{2}{*}{SFT based} & Qwen2.5-Coder-7B-Instruct & 56.36 & 57.54 & 43.89 & & 56.51 & 60.80 & 36.54 & & 56.95 & 63.83 & 38.44  \\
        & Qwen2.5-Coder-14B-Instruct & 65.87 & 66.24 & 62.37 & & 72.21	& 67.15	& 53.65 & & 70.15 & 75.49 & 56.19 \\
        \midrule
        \multirow{2}{*}{RL based} & \textbf{RuCo-C} (7B, BIRD trainset) & 
        68.15 & \textit{\underline{68.07}} & 67.33 & & 72.40 & 68.29 & 54.04 & & 75.60 & \textbf{77.84} & \textit{\underline{63.12}} \\
        & \textbf{RuCo-C} (14B, BIRD trainset) &  67.40 & 67.82 & 63.69 & &  \textbf{72.56} & \textbf{75.78} & \textbf{56.12} & & 70.08 & 77.41 & 56.29 \\
         \bottomrule
    \end{tabular}
    }
    \caption{Performance comparison of various methods across three testing datasets: \textbf{best results} are highlighted in bold, and \textit{\underline{second-best}} results are indicated in underline.}
    \vspace{-15pt}
    \label{tab:results}
\end{table*}

For static PR, we first consider whether the binary classification result predicted by the critique model is consistent with the ground-truth binary classification result $y$, which is $R_{out} = \mathbb{I}(R_{rubric}<1) \oplus y$, the symbol $\oplus$ means the XOR gate. When they are consistent (i.e. $R_{out}=1$), a coefficient value matching the binary classification result is assigned. Otherwise refer to the situation of $R_{out}=0$. To combat ``reward hacking'' caused by noisy labels in the dataset, where the model is unjustly penalized for sound reasoning due to conflicting flawed labels, we introduces a verify code (VC) mechanism to assess the credibility of intermediate reasoning, which the coefficent value denoted as $R_{cons}$. The whole static PR coefficient can be calculated as:
\begin{equation}
    \begin{aligned}
    \gamma_{s} &= \left\{
        \begin{aligned}
        & 2 * R_{rubric},  & \quad R_{out} = 1   \\
        & R_{cons}, \quad & R_{out} = 0 \\
        & 0, \quad & \text{otherwise}
        \end{aligned}
    \right. 
    \end{aligned}
\end{equation}

For cases where there is inconsistency with the final result, we further evaluate whether the reasoning process identifies errors. If an error is identified and the corrected SQL generated by the critique model passes verification, a small reward is given. Conversely, if an error is identified but the corrected SQL fails verification, which indicates low credibility in the critique model's reasoning process, then a small penalty coefficient is applied. The detailed design is formulated as:
\begin{equation}
    \begin{aligned}
        R_{cons} & = \left\{
        \begin{aligned}
            & 1 , & \mathbb{I}(R_{rubric}<1) = 1 \ \& \ R_{verify} = 1 \\
            & \text{-}1, & \mathbb{I}(R_{rubric}<1) = 1 \ \& \ R_{verify} = 0 \\
            &0, & \text{otherwise}
        \end{aligned}
        \right.
    \end{aligned}
\end{equation}

Here the $R_{verify} \in \{{0,1}\}$ means the corrected SQL fail or pass the verification, we use Test-Suite as the verifier by comparing between the predicted SQL and the corrected SQL. By validating the consistency of the generated correction SQL and the SQL being judged, we can provide positive feedback for sound reasoning despite label noise, exploring whether this mechanism can reduce the adverse effects of noisy data.

For dynamic PR, it is important to note that the total number of steps in the reasoning process may vary across different questions. Based on our statistical data, the number of steps in the reasoning process exhibits a clear correlation with question difficulty. Specifically, questions of easy and medium difficulty typically involve around 5 reasoning steps, while those of hard and extra difficulty generally require 6 to 7 or more steps. Given this data characteristic, we further introduce a dynamically adjusted coefficient $\gamma_{d}$ for the rubric-based process reward, which is defined based on the number of reasoning steps as follows:
\begin{equation}
    \begin{aligned}
         \gamma_{d} & = \left\{ 
         \begin{aligned}
             & 1, & N_{i} > 5 \\
             & 0, & \text{otherwise}
         \end{aligned}
         \right.
    \end{aligned}
\end{equation}
Hence, the final total reward score can be formulated as:
\begin{equation}
\label{eq:final_reward_score}
    R_{total} = R_{format} + 2 * R_{out} + (\gamma_{s} + \gamma_{d}) * R_{rubric}   \\
\end{equation}

\section{Experiment}

\subsection{Experimental Setup}

\noindent \textbf{Dataset.} Our experiments are validated in three public text-to-SQL datasets: Spider~\cite{yu2018spider}, BIRD~\cite{li2023can} and Spider-DK~\cite{gan2021exploring}. Since the information contained in these development datasets does not align with the requirements of the evaluation scenario, we constructed additional negative sample data based on them. The detailed construction process can be seen in Appendix~\ref{sec:dataset_construction}. We conducted statistical analyses on the size and question difficulty distribution of the three datasets, with detailed information provided in Table~\ref{tab:testset statistic}.

\noindent \textbf{Baselines.} We compared open-source and closed-source LLMs as baselines, all of which were tested via the Prompt Engineering (PE) approach. For reproducibility, the prompts used in these baseline methods are identical to those in our model’s training and inference, please refer to the section~\ref{sec:prompt_inference} in Appendix for details. Among proprietary models, we selected OpenAI’s state-of-the-art o3-mini and GPT-4.1. As for open-source models, we chose large foundational models (e.g., DeepSeek-R1~\cite{deepseekai2025deepseekr1incentivizingreasoningcapability}, DeepSeek-V3~\cite{DBLP:journals/corr/abs-2412-19437} and Kimi-K2~\cite{kimiteam2025kimik2openagentic}) based on scale, plus smaller-scale models (e.g., about 7B scale model DeepSeek-Coder-6.7B-Instruct and Qwen2.5-Coder-7B-Instruct, and 14B scale model Qwen2.5-Coder-14B-Instruct) for comparison. Additionally, we compared the performance of models trained with SFT, and further evaluated the model's effectiveness with and without the critique responses for its outputs.

\noindent \textbf{Evaluation Metrics.} For evaluating the correctness of text-to-SQL, we employ metrics such as \textit{AUC}, \textit{accuracy (ACC)}, and \textit{F1-score (F1)}. Considering that the classes are imbalanced in our testing dataset, we adopt AUC metric to measure the model's ability of distinguishing correct and wrong predictions, and the F1-score is used to balance the trade-off between avoiding false positives and capturing true positives. To offer a straightforward overall performance, the ACC is used to represent the proportion of correct predictions. To facilitate result comparison, all numerical results in this paper are presented in percentage form.


\smallskip
\noindent \textbf{Implementation Details.} The SFT and RL experiments were performed based on 8 GPUs. For RL training, we employed the GRPO strategy~\cite{shao2024deepseekmathpushinglimitsmathematical} based on the verl architecture~\cite{sheng2025hybridflow}, with the following hyperparameters configurations: a batch size of 32, a learning rate of $1e^{-6}$, and 5 rollouts. Additional details about hyperparameters are provided in Appendix~\ref{sec:ablation}.

\begin{table*}[th!]
\centering
\setlength{\tabcolsep}{6pt}
\renewcommand{\arraystretch}{1.0}
    \resizebox{\textwidth}{!}{
        \begin{tabular}{l cccccc c cccccc c cccccc}
            \toprule
            \multirow{2}{*}{\textbf{Method}} & \multicolumn{6}{c}{\textbf{Spider}} & &  \multicolumn{6}{c}{\textbf{BIRD}} & & \multicolumn{6}{c}{\textbf{Spider-DK}} \\
            \cmidrule(lr){2-7} \cmidrule(lr){9-14} \cmidrule(lr){16-21}
            & \textbf{AUC} & & \textbf{ACC} & & \textbf{F1} & & &  \textbf{AUC} & & \textbf{ACC} & & \textbf{F1} & & &\textbf{AUC} & & \textbf{ACC} & & \textbf{F1} & \\
            \midrule
            EX (w/o rubric) & 58.73 & &  59.31 & & 52.92 & & &  67.10  & & 66.17 & & 48.65 & & & 59.00 &  & 67.55 & & 40.12 & \\
            EX & 65.19 & \textcolor{red}{$\uparrow$ 6.36} & 66.06 & \textcolor{red}{$\uparrow$ 6.75} & 58.05 & \textcolor{red}{$\uparrow$ 5.13} & & 66.88 & \textcolor{gray}{$\downarrow$ 0.22}& \textbf{75.01} & \textcolor{red}{$\uparrow$ 8.84}& 49.04 & \textcolor{red}{$\uparrow$ 0.39} & & 69.59 & \textcolor{red}{$\uparrow$ 10.59} & \textbf{79.01} & \textcolor{red}{$\uparrow$ 11.46} & 55.73 & \textcolor{red}{$\uparrow$ 15.61} \\
            \midrule
            EX+PR (static) & 65.40 & \textcolor{red}{$\uparrow$ 6.57} & 64.90 & \textcolor{red}{$\uparrow$ 5.59}  & 67.12 & \textcolor{red}{$\uparrow$ 14.20}  & & 68.97  & \textcolor{red}{$\uparrow$ 1.87} & 62.71 &\textcolor{gray}{$\downarrow$ 4.54} & 50.18 &\textcolor{red}{$\uparrow$ 1.53 }& & 74.92 &\textcolor{red}{$\uparrow$ 15.92} & 74.16 & \textcolor{red}{$\uparrow$ 6.61} & 61.35 & \textcolor{red}{$\uparrow$ 21.23} \\
            
            EX+RR+VC (static) & 66.36 & \textcolor{red}{$\uparrow$ 7.63} & 66.12 & \textcolor{red}{$\uparrow$ 6.81}& 66.34 & \textcolor{red}{$\uparrow$ 13.42} &  &70.58 & \textcolor{red}{$\uparrow$ 3.48} & 69.97 & \textcolor{red}{$\uparrow$ 3.80}& 52.65 & \textcolor{red}{$\uparrow$ 4.00} & & 75.46 & \textcolor{red}{$\uparrow$ 16.46}& 77.73 & \textcolor{red}{$\uparrow$ 10.18}& 62.94 & \textcolor{red}{$\uparrow$ 22.82}\\
             \midrule
            EX+PR (static+dynamic) & 67.34 &\textcolor{red}{$\uparrow$ 8.61} & 67.46 & \textcolor{red}{$\uparrow$ 8.15} &	65.42 &\textcolor{red}{$\uparrow$ 12.50} & & 70.83	& \textcolor{red}{$\uparrow$ 3.73}& 72.35	& \textcolor{red}{$\uparrow$ 6.18} & 53.37 & \textcolor{red}{$\uparrow$ 4.72}& & 74.90 & \textcolor{red}{$\uparrow$ 15.90} & 78.48 & \textcolor{red}{$\uparrow$ 10.93}&	62.59  & \textcolor{red}{$\uparrow$ 22.47}\\
            EX+PR+VC (static+dynamic) & \textbf{68.15} & \textcolor{red}{$\uparrow$ 9.42} &\textbf{68.07} & \textcolor{red}{$\uparrow$ 8.76}& \textbf{67.33} & \textcolor{red}{$\uparrow$ 14.41} & &\textbf{72.40} & \textcolor{red}{$\uparrow$ 5.20}& 68.29 & \textcolor{red}{$\uparrow$ 2.12}& \textbf{54.04} & \textcolor{red}{$\uparrow$ 5.39}& & \textbf{75.60} & \textcolor{red}{$\uparrow$ 16.60}& 77.84 & \textcolor{red}{$\uparrow$ 10.29} & \textbf{63.12} & \textcolor{red}{$\uparrow$ 23.00}\\
            \bottomrule
        \end{tabular}
    }
    \caption{Ablation study results on RL reward function designing across three testing datasets, all variants use Qwen2.5-Coder-7B-Instruction as the backbone model. \textbf{Best results} are highlighted in bold, with accompanying performance improvements or declines calculated relative to the EX (w/o rubric) method.}
    \label{tab:ablation}
    \vspace{-15pt}
\end{table*}

\vspace{-3pt}
\subsection{Performance Evaluation}

We conduct a comparative analysis of our proposed method \modelname~against the performance of baselines. To evaluate generalization capability, our \modelname~model was trained on only a subset of the BIRD training dataset during the GRPO phase. Evaluation results across three test datasets are presented in Table~\ref{tab:results}, from which the following key findings are derived: i) Both open-source and proprietary LLMs exhibit significantly superior evaluation performance across all three test datasets compared to small-scale models with 7B or 14B parameters. ii) For models trained via SFT, where training was restricted to a subset of the Spider dataset, cross-scale comparisons of model parameters reveal that 14B-parameter models yield more pronounced performance gains from SFT than their 7B-parameter counterparts. iii) Our proposed model \modelname, which integrates the custom-designed reward function into RL training, achieves notably better evaluation performance than SFT-trained models. Even when RL training is confined to a subset of the BIRD dataset, the model still demonstrates substantial improvements in inference performance on the other two test datasets.

\subsection{Ablation Study}
To verify the rationality and effectiveness of the reward design, we conducted a series of ablation studies of GRPO methods with varied reward signals, i.e., EX, EX+PR and EX+PR+VC (\modelname), the detailed description can be seen in Appendix~\ref{sec:ablation}.

The performance of the aforementioned variant methods is presented in Table~\ref{tab:ablation}. Analysis of these results yields the following conclusions. First, the two methods at the top of the table are both EX variants, differing in whether rubric information is output in the evaluation results during the SFT phrase. A comparison of their results demonstrates that incorporating rubric-based critique responses into SFT training achieves superior effectiveness. 
Second, comparisons between the two types of methods (EX+PR and EX+PR+VC) in the middle and bottom sections of the table confirm the validity of integrating the verify code strategy for overall performance. This effectiveness is particularly notable under static coefficients, with an average $4\%$ improvement observed in the ACC metric.
Finally, overall comparisons indicate that adding our designed process reward to the judge model significantly enhances evaluation performance, compared to judge models that use only EX as the reward. Furthermore, we integrate process reward with both static and dynamic coefficients, which further improves the model’s overall evaluation capability.

\subsection{Reward Score Consistency Analysis}

\begin{figure}
    \centering
    \begin{subfigure}[b]{0.45\linewidth}
        \includegraphics[width=\linewidth]{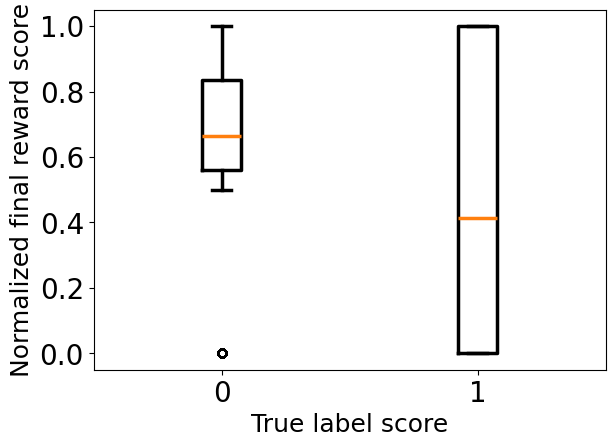}
        \caption{}
        \label{fig:reward_score_dist_RL_w_EX_box}
    \end{subfigure}
    \hfill
    \begin{subfigure}[b]{0.45\linewidth}
        \includegraphics[width=\linewidth]{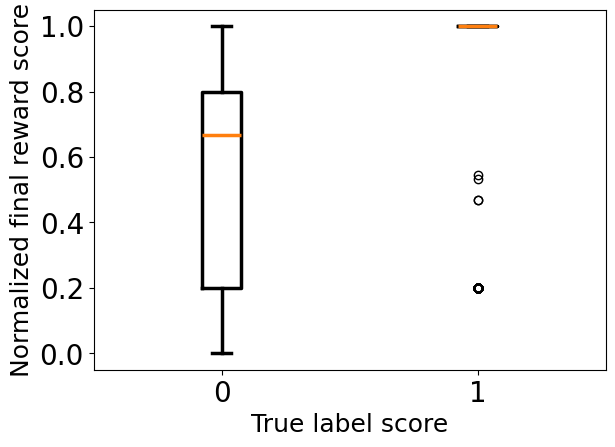}
        \caption{}
        \label{fig:reward_score_dist_RL_w_EX_PR_box}
    \end{subfigure}
    
    \vspace{10pt}
    \begin{subfigure}[b]{0.45\linewidth}
        \includegraphics[width=\linewidth]{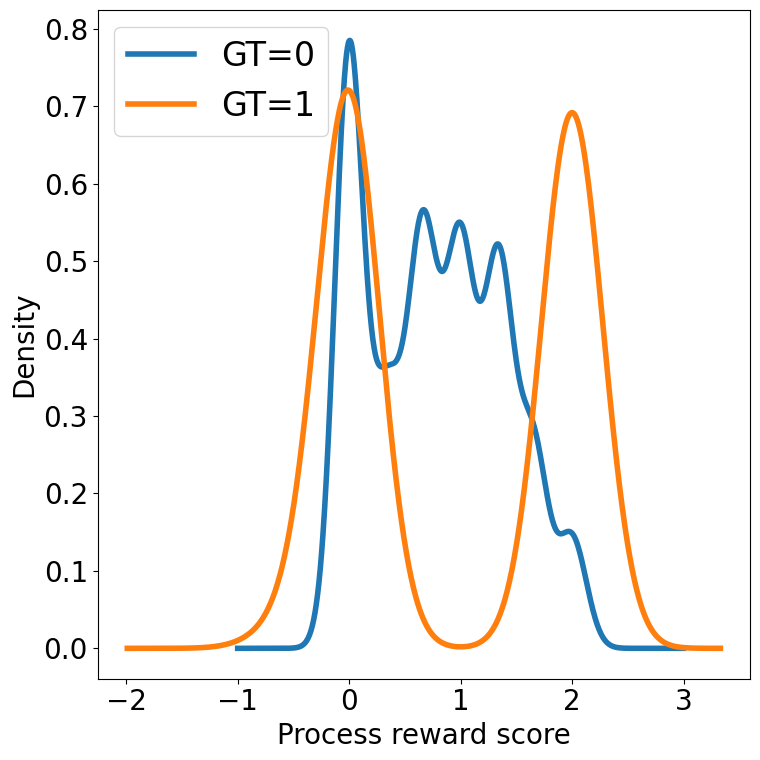}
        \caption{}
        \label{fig:reward_score_dist_RL_w_EX_kde}
    \end{subfigure}
    \hfill
    \begin{subfigure}[b]{0.45\linewidth}
        \includegraphics[width=\linewidth]{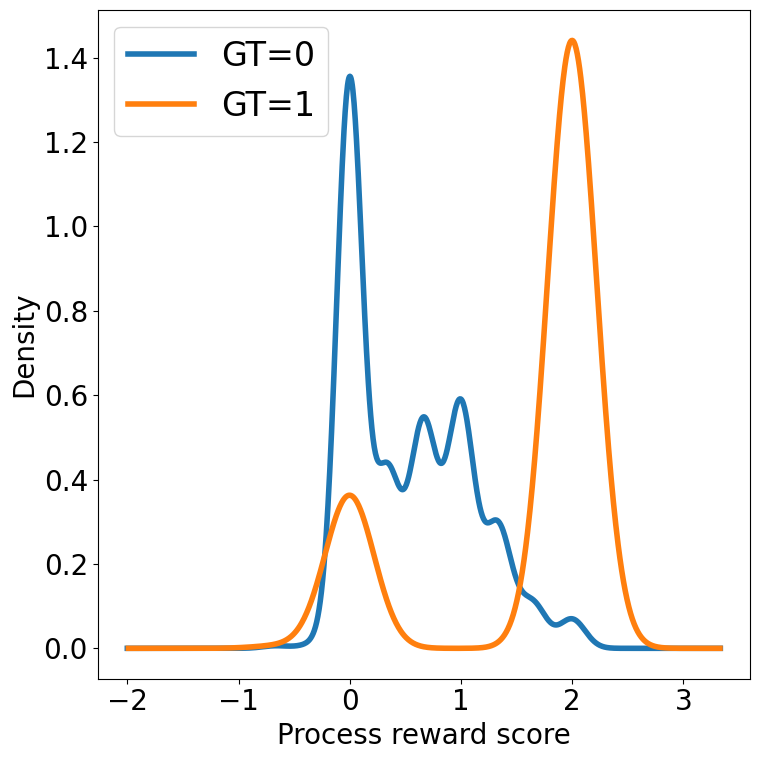}
        \caption{}
        \label{fig:reward_score_dist_RL_w_EX_PR_kde}
    \end{subfigure}
    \caption{Box and kernel density plots comparing the distributions of ground truth labels and reward function under the EX (left) and our \modelname~method (right).}
    \vspace{-10pt}
    \label{fig:reward_score_dist_cmp}
\end{figure}

To more intuitively illustrate the relationship between the designed reward scores and ground truth labels, we used box plots and kernel density plots to compare the distributions of reward scores and true labels obtained by EX method and \modelname~method. As shown in Figure~\ref{fig:reward_score_dist_cmp}, box plots (a) and (b) display the differences in the central tendency of the normalized final reward scores for positive and negative samples, while kernel density plots (c) and (d) show the probability density distributions of process reward scores across positive (GT=1) and negative sample (GT=0) categories.

It can be clearly observed that, compared to plot (a), the distribution of reward scores for negative samples in plot (b) is more concentrated. Correspondingly, from the comparison of the kernel density plots, the distributions of the two categories in plot (c) overlap considerably, whereas the overlap between the two distributions in plot (d) is significantly lower. This indicates that \modelname~method outperforms EX method in terms of classification effectiveness, and it verifies that the inclusion of process reward scores significantly improves the final classification performance and notably reduces the proportion of false negatives.

\subsection{Case Study}

As a query-based comparative assessment method, EX scoring may yield false judgments in specific scenarios. Firstly, inaccuracies in the gold SQL can directly lead to misclassifications. Secondly, most benchmarks offer just one reference answer leading it struggles to identify alternative solutions. Lastly, incomplete database values cause EX scoring to fail in conducting comprehensive and accurate assessments, resulting in false outcomes.

Here, we give a case in Figure~\ref{fig:case_1} to illustrate that the execution result does not reflect the order required in the user's question. Additionally, the incorrect gold SQL leads to an EX scoring error, while the generated critique response via our proposed \modelname~can identify this issue in the pred SQL without seeing the gold SQL. Specifically, the critique response excerpt presented herein clearly identifies that the formulation of the predicted SQL fails to align with the intent of ``first paid customer'' specified in the user’s query. Furthermore, the operational recommendations outlined in the critique response indicate that the transaction table should be sorted in ascending order (ASC) by the time field. In contrast, the gold SQL employs a descending order (DESC) for this sorting operation, this discrepancy directly reveals the inaccuracy of the gold SQL. Reasoning based on the critique response, our model ultimately obtained a correct assessment. Further case analyses can be found in the appendix~\ref{appendix:case analysis}.

\begin{figure}
    \begin{minipage}[t]{0.49\textwidth}
        \begin{tcolorbox}[tinycasebox={Order-insensitive \& Wrong Gold SQL}]{
            \vspace{-5pt}
            \textbf{Question:} Which country's gas station had the \colorbox{yellow!30}{first paid customer} in 2012/8/25? \\
            \textbf{Pred SQL:} \\
            \vspace{-10pt}
            \begin{lstlisting}[style=tinysqlstyle, label=case_1:pred_sql]
SELECT g.country
FROM transactions_1k t
JOIN gasstations g ON t.gasstationid = g.gasstationid
WHERE t.date = '2012-08-25'
LIMIT 1 \end{lstlisting}
            \textbf{Gold SQL:} \\
            \vspace{-10pt}
            \begin{lstlisting}[style=tinysqlstyle, label=case_1:gt_sql]
SELECT T2.Country 
FROM transactions_1k AS T1 
INNER JOIN gasstations AS T2 
ON T1.GasStationID = T2.GasStationID 
WHERE T1.Date = '2012-08-25' 
ORDER BY T1.Time DESC 
LIMIT 1 \end{lstlisting}
            \textbf{Critique Response (part):} \\
            \colorbox{yellow!30}{\parbox {\dimexpr \linewidth-2\fboxsep\relax}{
            \textbf{<think>}\\
            ...\\
            4: Did I correctly identify the first paid customer? \\
            - The query uses LIMIT 1 to get the first transaction matching the date filter \textcolor{red}{but doesn’t guarantee it’s the first paid customer by time}. To do so, transactions should be ordered by time \textcolor{red}{ascendingly}.\\
            \textbf{</think>} \\
            \textbf{<result>} False \textbf{</result>}\\
            ...
                }
            }
        }
        \end{tcolorbox}
    \end{minipage}
    \caption{The false positive example of EX scoring (EX score is 1 while the true label is 0), along with the generated critique response by our \modelname~to indicate the correct judgment.}
    \label{fig:case_1}
    \vspace{-15pt}
\end{figure}

\section{Conclusion}
In this paper, we propose a novel generative judge model \textbf{\modelname}~tailored for text-to-SQL evaluation. By integrating interpretable rubrics with the designed consistent reward functions, \modelname~enables fine-grained, query-specific automatic evaluation, thereby eliminating the need for human intervention and addressing key bottlenecks in the text-to-SQL assessment field. Complementary experiments conducted on three distinct test sets demonstrate that our proposed model achieves superior evaluation performance compared to existing baseline methods. Looking ahead, we plan to explore the integration of \modelname~into end-to-end text-to-SQL training pipelines, enabling real-time feedback during the post-training phrase to accelerate performance improvement.

\section{Limitations}

Despite the proposed evaluation method eliminating reliance on SQL query comparisons, avoiding gold SQL annotation costs, and achieving better performance than existing baseline methods, it still has limitations. Specifically, the small size of the training dataset restricts the generalization ability of the post-training model, leaving room for further improvement.  Accordingly, future work will explore integrating this evaluation method into training data synthesis and text-to-SQL model training. Generated results will be fed back into the training and capability enhancement of the evaluation model, ultimately forming a complete closed loop of data generation, training, and evaluation.

\bibliography{acl_latex}

\clearpage

\newpage
\twocolumn

\appendix

\section{Dataset Construction}
\label{sec:dataset_construction}
In this chapter, we elaborate on the construction methods of the training sets and test sets employed in the experiments.

\begin{table*}
\centering
\renewcommand{\arraystretch}{1.0}
    \resizebox{0.7\textwidth}{!}{
        \begin{tabular}{l l ccc}
            \toprule
          \textbf{Distribution} & \textbf{Label} & \textbf{Spider} & \textbf{BIRD} & \textbf{Spider-DK} \\ 
             \midrule
            \multirow{3}{*}{\textbf{Binary Distribution}} & Total & 1644 & 2977 & 1877 \\
            & Positive & 776 (47.20\%) & 693 (23.28\%) & 503 (26.80\%) \\
            & Negative & 868 (52.80\%) & 2284 (76.72\% ) & 1374 (73.20\%) \\
            \midrule
            \multirow{4}{*}{\textbf{Hardness Distribution}} & Easy & 57 (3.47\%) & 18 (0.60\%) & 79 (4.21\%) \\
            & Medium & 620 (37.71\%) & 1047 (35.17\%) & 1017 (54.18\%) \\
            & Hard & 246 (14.96\%) & 811 (27.24\%) & 247 (13.16\%) \\
            & Extra & 721 (43.86\%) & 1101 (36.98\%) & 534 (28.45\%)\\         
            \bottomrule
        \end{tabular}
    }
    \caption{Statistical overview of binary and difficulty distributions across three testing datasets.}
    \label{tab:testset statistic}
\end{table*}

\subsection{Training Dataset}
The training data is primarily divided into two phases: the SFT phase and the RL phase. For the SFT phase, we mainly conduct data synthesis based on the Spider training data, whereas for the RL phase, we construct data using the more complex BIRD training data.

The original Spider~\cite{yu2018spider} dataset contains 10,181 questions and 5,693 complex SQL queries, sourced from 200 multi-table databases across 138 domains. The entire dataset is divided into three non-overlapping subsets: training (8,659), development (1,034) and test datasets. Since the original Spider dataset includes only gold SQL queries, our experiments require a large number of erroneous SQL queries as negative samples. To address this, we adopt a synthesized dataset~\cite{chen2023text} with erroneous SQL queries based on the the training set of Spider: this dataset retains the question information from the original Spider and generates multiple erroneous SQL queries for each question. Using this dataset, we constructed approximately 40,000 data samples and generated critique responses for both positive and negative samples via the proposed multi-agent framework. Considering the need for balanced positive-negative sample ratios, we ultimately sampled and used 15,000 of these samples as the training dataset for the SFT.

The original BIRD~\cite{li2023can} is another large-scale and cross-domain text-to-SQL dataset known for its complexity. It contains 9,428 training examples and 1,534 development examples, which drawn from 95 databases cross more than 37 professional fields, making it a challenging testbed for evaluating the generalization capability in text-to-SQL task. In RL training, we also need to construct a set of negative samples. To this end, we generate SQL queries using several text-to-SQL models, determine their correctness via existing evaluation methods (e.g., Test-Suite Accuracy), and filter out samples with erroneous SQL queries as negative samples. While we now have positive and negative samples along with their corresponding labels as ground truth for feedback signals, the BIRD dataset is widely recognized for its ``dirty'' characters: potentially incorrect SQL queries, ambiguous or poorly described column names, and databases containing null values or irregular encodings. Consequently, we sample and further validate the data, using a combination of multiple automated evaluation methods and limited manual checks to finalize whether each sample is positive or negative.

\subsection{Testing Dataset}

For the testing dataset, we constructed positive and negative samples following a similar approach to training dataset construction, using existing development sets as the basis. The statistical details of the final test sets employed in our experiments are presented in Table~\ref{tab:testset statistic}. In terms of binary classification distribution, we not only built a Spider test set with nearly balanced positive-negative ratios but also constructed BIRD and Spider-DK test sets with higher proportions of negative samples. This design aims to comprehensively test the evaluation model’s ability to detect erroneous SQL. Additionally, regarding the distribution of question difficulty, our test set composition prioritized evaluating the model’s performance on questions of moderate or higher difficulty.

\section{Implement Details}

\subsection{Baselines Implementation}
In this subsection, we illustrate the detailed implementation of the baselines used in our experiments. For open source and proprietary models, we utilize the same prompts (as shown in \ref{sec:prompt_inference}) that were employed in the training phase of our proposed model for inference to ensure consistency. For each test set, we perform three separate inference runs and average the results to yield the final performance. 

Regarding SFT training, data was sampled from the constructed training set as illustrated in Appendix~\ref{sec:dataset_construction} to maintain a balance between positive and negative samples; subsequently, the appropriate SFT checkpoint was selected as the initial policy model for the GRPO training. 

\begin{table*}
\centering
    \renewcommand{\arraystretch}{1.0}
    \begin{tabular}{l l}
        \toprule
        \textbf{Hyperparameter} & \textbf{Default Value} \\
        \midrule
        Optimizer & AdamW \\
        Learning Rate & $1e^{-6}$  \\
        Training Batch Size & 32 \\
        Training Total Epochs & 3 \\
        Max Prompt Length & 4096 \\
        Max Response Length & 2048 \\
        PPO Mini Batch Size & 8 \\
        Log Probability Micro Batch Size per GPU (Rollout) & 8 \\
        Tensor Model Parallel Size (Rollout) & 2 \\
        Group Sampling Number (Rollout) & 5 \\
        GPU Memory Utilization (Rollout) & 0.6 \\
        Temperature (Rollout) & 1.0 \\
        Do Sample (Rollout) & True \\
        Clip Ratio ($\epsilon$ in Eq.~(\ref{eq:opt_objective})) & 0.2 \\
        KL Loss Coefficient ($\beta$ in Eq.~(\ref{eq:opt_objective})) & 0.001 \\
        Normalize Advantages by Standard Deviation & True \\
        \bottomrule
    \end{tabular}
    \caption{Hyperparameters setting during GPRO training.}
    \label{tab:hyperparameter}
\end{table*}

\subsection{Ablation Studies Implementation}
\label{sec:ablation}
The ablation studies of the GRPO methods shown in Table~\ref{tab:ablation} are all based on Qwen2.5-Coder-7B-Instruct, which mainly include three types of variant, within each variant category, there are two distinct methods:

\begin{itemize}
    \item EX as a reward: the reward score used in GPRO consists primarily of a format score (check the format of critique response) and a correction score (check the binary result is consistent with the ground-truth label). 
    \begin{itemize}
        \item EX (w/o rubric): during the SFT training phrase, the training examples do not include any rubric-related critique response but only binary results.
        \item EX: during the SFT training phrase, the training examples contain the critique response as in \modelname.
    \end{itemize}
    \item EX+PR as a reward: Based on using EX as reward, this variant further checks the QA pairs (i.e., content in tag <think> of critique response) generated by the judge model to determine whether the answer in each QA pair is correctly elaborated, i.e., whether the elaboration aligns with the intentions of the query and the given information.
    \begin{itemize}
        \item EX+PR (static): based on the coefficient $\gamma_{s}$ but without considering the situation $R_{out}=0$ .
        \item EX+PR (static+dynamic): based on the EX+PR (static) variant, further consider the coefficient $\gamma_{d}$ to dynamically combine the rubric-guided process reward $R_{rubric}$.
    \end{itemize}
    \item EX+PR+VC as a reward: Based on EX+PR, we further verify the credibility of the judge information through the verify code (VC) method: by re-judging the generated correction SQL, if the correction SQL is correct, the PR credibility is considered positive; otherwise, it is negative. 
    \begin{itemize}
        \item EX+PR+VC (static): only using coefficient $\gamma_{s}$ to combine the rubric-guided process reward $R_{rubric}$. 
        \item EX+PR+VC (static+dynamic): using the final total reward score as shown in Eq.~(\ref{eq:final_reward_score}).
    \end{itemize}
\end{itemize}

All the aforementioned variant methods are trained based on the verl architecture~\cite{sheng2025hybridflow} and share largely consistent core experimental hyperparameters, with specific details provided in Table~\ref{tab:hyperparameter}. There are some different settings compared to the original GRPO paper, for example, we adopt the ``token-mean'' configuration for loss aggregation instead of sample-level loss which may be unstable in long-CoT scenarios.

\begin{figure*}[th!]
    \centering
    \begin{subfigure}[b]{0.45\linewidth}
        \includegraphics[width=\linewidth]{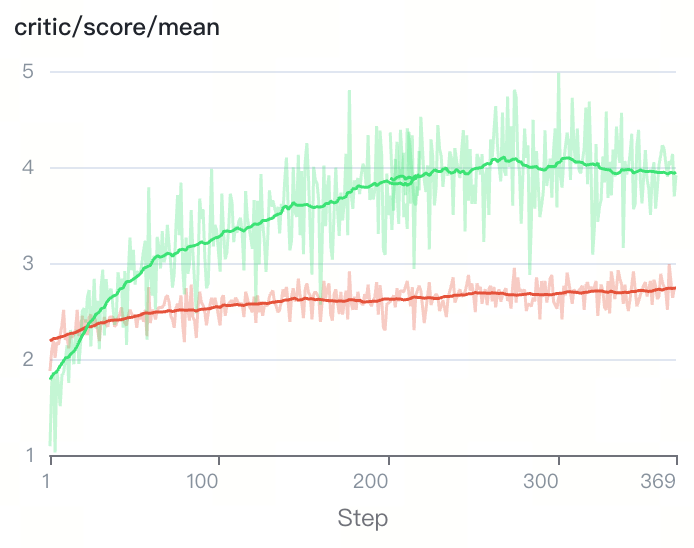}
        \caption{Average reward score for batched training data}
        \label{fig:reward_score_train}
    \end{subfigure}
    \hfill
    \begin{subfigure}[b]{0.45\linewidth}
        \includegraphics[width=\linewidth]{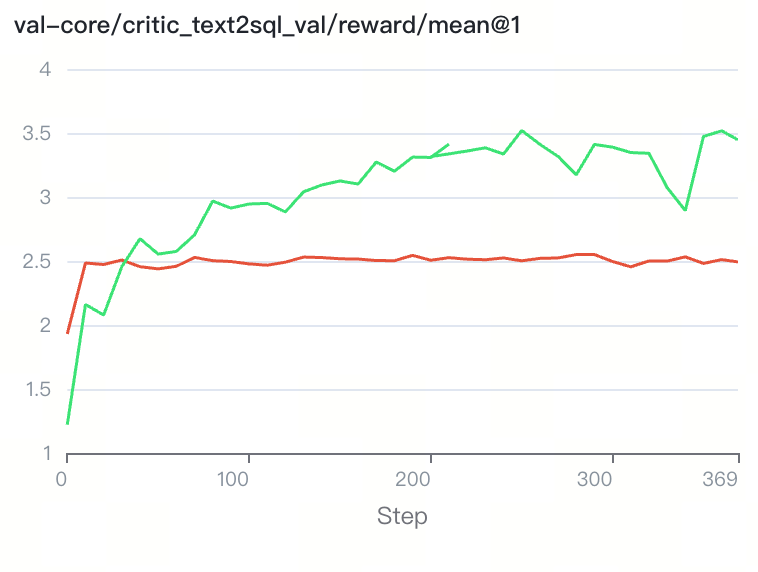}
        \caption{Reward score of validation set}
        \label{fig:reward_score_val}
    \end{subfigure}
    \caption{The reward score trends on the training set~(left) and validation set~(right) during GRPO training steps. The red curve corresponds to baseline model based on the EX reward, while the green curve indicates to our \modelname~model using designed reward function. Both model are training with Qwen2.5-Coder-7B-Instruct backbone under the GRPO framework. Despite starting from a relatively low initial score, our \modelname~model exhibits a distinct upward trend, indicating that the feedback obtained with increasing training steps yields significant optimization for the model.}
    \label{fig:reward_score_cmp}
\end{figure*}

\subsection{Visualized Comparison on Training Process}

\begin{figure*}[th!]
    \centering
    \begin{subfigure}[b]{0.45\linewidth}
        \includegraphics[width=\linewidth]{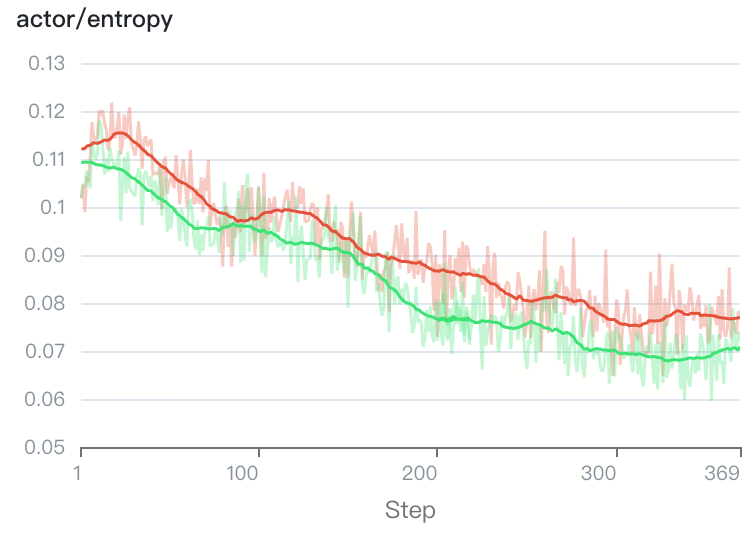}
        \caption{The entropy loss under training steps}
        \label{fig:entropy_cmp}
    \end{subfigure}
    \hfill
    \begin{subfigure}[b]{0.45\linewidth}
        \includegraphics[width=\linewidth]{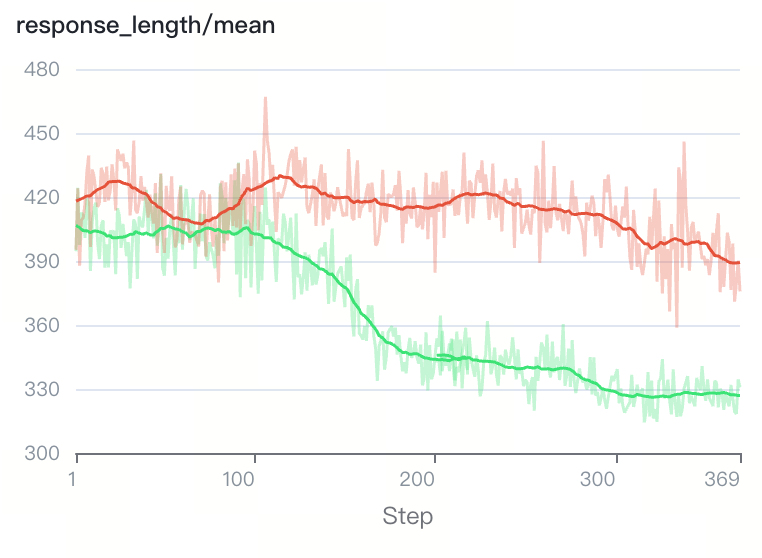}
        \caption{The averaged response length under training steps}
        \label{fig:response_len_cmp}
    \end{subfigure}
    \caption{The trend of entropy loss and averaged response length during GRPO training steps. The red curve corresponse to baseline model based on the EX reward, while the green curve is our \modelname~model using designed reward function. Both model are training with Qwen2.5-Coder-7B-Instruct backbone under the GRPO framework.}
    \label{fig:entropy_response_len_cmp}
\end{figure*}

To visually examine the GPRO process under different reward functions, we present the variation trends of reward scores across training steps (Figure~\ref{fig:reward_score_cmp}), along with changes in entropy loss and generated response lengths during training (Figure~\ref{fig:entropy_response_len_cmp}). In this two figures, the red curve corresponds to the GPRO method using only EX rewards, while the green curve represents our proposed method \modelname~integrating process and outcome rewards.

As shown in Figure~\ref{fig:reward_score_cmp}, the left and right panels depict the trend of average reward scores across training steps and reward score changes on the validation set, respectively. 
It can be observed that although the reward score of our method was lower than that of the EX method at the initial training stage, as training iterations progressed, the model captured more feedback signals, guiding it toward optimization for better reward objectives. Both the training and validation sets exhibit a distinct upward trend in reward scores. In contrast, the red curve (EX method) shows limited growth in reward scores. This may stem from the discrete distribution of EX-based reward scores, which fails to reflect fine-grained feedback signals for individual samples, thereby restricting its ability to guide model optimization.
Despite the differences in the numerical ranges of the reward scores that our method shows a wider range and higher values in the growth trend of the training set, the results of the validation set clearly indicate that the EX method reaches a plateau early in training without further improvement. This confirms that the performance of our method continues to improve throughout the training process.

Figure~\ref{fig:entropy_cmp} illustrates the trend of entropy loss in the policy model during RL training for both methods. In reinforcement learning, an action that receives both high/low probability and high/low advantage will lower the entropy, and vice versa. According to the entropy mechanism~\cite{cui2025entropymechanismreinforcementlearning}: at the beginning stage, the policy exhibits a high covariance with the training data, indicating that its confidence is well calibrated. This allows it to safely exploit high-confidence trajectories, strengthening its beliefs and minimizing entropy.
The trends of the two curves in Figure~\ref{fig:entropy_cmp} generally confirm the effectiveness of both methods in GPRO training. Comparing their values, the entropy loss of our model \modelname~consistently remains below that of the EX method, indicating that our method has higher confidence in exploration. Additionally, based on the negative exponential relationship between performance and entropy, this indirectly demonstrates that our method outperforms the EX method in terms of performance.

Furthermore, we also analyzed the length of the responses generated by the model during training, with relevant results presented in Figure~\ref{fig:response_len_cmp}. It can be observed that between training steps 100 and 200, the response length generated by our model showed a significant reduction compared to the baseline model, and then stabilized around 300 steps. Given that the response comprises three components, the primary reduction in length is likely concentrated in the rubric-related reasoning phase. This indicates that as confidence in training exploration increases, the rubric reasoning information becomes more explicit and focused. By eliminating irrelevant rubric-related questions, the model can converge more quickly to core evaluation criteria while maintaining its evaluation capability.

\subsection{Performance Comparison Group By Hardness}

\begin{figure*}[th!]
    \centering
    \begin{subfigure}[b]{0.45\linewidth}
        \includegraphics[width=\linewidth]{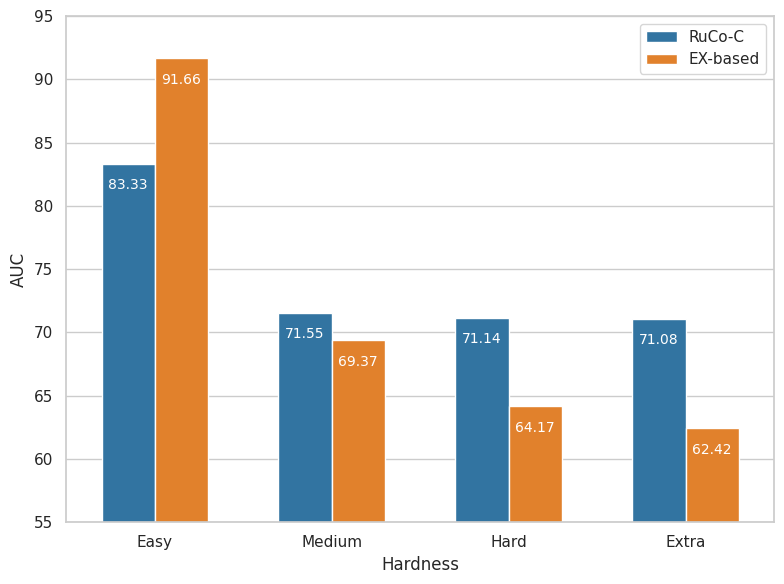}
        \caption{Performance comparison under AUC metric}
    \end{subfigure}
    \hfill
    \begin{subfigure}[b]{0.45\linewidth}
        \includegraphics[width=\linewidth]{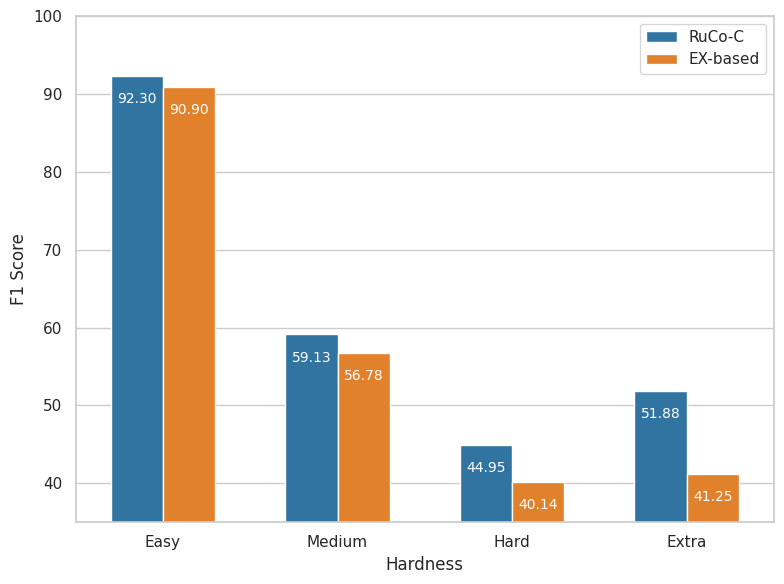}
       \caption{Performance comparison under F1-score metric}
    \end{subfigure}
    \caption{Performance comparison between our \modelname~and baseline methods on AUC and F1- score metrics across different hardness of questions in BIRD test set.}
    \label{fig:hardness_cmp}
\end{figure*}

Beyond the overall performance comparison, we further explore the performance differences of the models across varying difficulty levels of the questions (named hardness here). As shown in Figure ~\ref{fig:hardness_cmp}, we compare our \modelname~model with the EX-based reward model under different hardness in the BIRD test set. As indicated in Table~\ref{tab:testset statistic}, the BIRD test set contains fewer easy questions, most of which are medium or higher difficulty. Given the imbalanced positive-negative sample ratio in the BIRD test set, we use the AUC metric for comparison. It can be observed that the EX-based method performs better on simple questions, but our method demonstrates a significant advantage on moderate and higher difficulty levels, which account for a larger proportion of the dataset. Additionally, our method achieves better results in terms of F1-score, indicating that it also exhibits superior performance in avoiding false positives compared to the EX-based method.

\section{Examples of False Annotations Based on EX}
\label{appendix:case analysis}

As a pairwise assessment approach, EX scoring has the potential to yield inaccurate outcomes in specific scenarios. Firstly, inaccuracies in gold SQL which can directly lead to misclassifications, wrongly identifying correct results as false positives. Secondly, dealing with multiple reference answers poses a challenge.  Most benchmarks offer just one reference answer. As a result, EX scoring, constrained by narrow criteria, struggles to identify valid alternative solutions, incorrectly classifying them as false negatives. Lastly, incomplete database values contribute to misjudgments. Without sufficient data for evaluation, EX scoring cannot conduct comprehensive and accurate assessments, resulting in false outcomes.

As illustrated in Figure~\ref{fig:case2}, here is an example of a false negative due to the inconsistency between the given input and the gold SQL. In both the user's question and the evidence, the term is stated as ``cryokinesis'' (with a lowercase initial letter). However, the value corresponding to the ``power\_name'' field in the actual database is ``Cryokinesis'' (with an initial letter in uppercase). After executing the SQL statements, the gold SQL can retrieve the result, while the predicted SQL returns an empty result, which leads to a false negative in the EX score. However, based on the contextual information given, the predicted SQL is correct, as it is derived from step-by-step reasoning in the QA process without knowing the complete values in the database. 

Another example (as shown in Figure~\ref{fig:case3}) involves cases where there are field-value errors in the predicted SQL, yet its execution results are consistent with those of the gold SQL. Specifically, the user's query requires counting players involved in attack actions. The filter condition ``defensive\_work\_rate=`low''' in the gold SQL only emphasizes that the players have a low probability of participating in defense, without indicating a high probability of participating in attacking. Additionally, the predicted SQL contains an error in the value of `attack\_work\_rate' in its filter condition, which should be `high' instead of `defensive'. Such a value error evaded the executor's error detection capability due to the logical condition `!='. However, in the critique response generated by our model, the red-highlighted text shows that our model can identify such errors and provide correct modification suggestions.

\begin{figure*}
    \begin{minipage}[t]{\textwidth}
        \begin{tcolorbox}[casebox={Input and Gold SQL Inconsistency}]
            \rndlabel{Question} What is the power ID of \colorbox{yellow!30}{cryokinesis}? \\[5pt]
            \rndlabel{Evidence} power ID refers to superpower.id, cryokinesis refers to power\_name = `cryokinesis' \\[5pt]
            \rndlabel{Pred SQL} 
            \begin{lstlisting}[style=sqlstyle, label=case2:pred_sql]
SELECT id 
FROM superpower 
WHERE power_name = 'cryokinesis'    \end{lstlisting} 
            \rndlabel{Gold SQL} 
            \begin{lstlisting}[style=sqlstyle, label=case2:gt_sql]
SELECT id 
FROM superpower 
WHERE power_name = 'Cryokinesis' \end{lstlisting}
            \rndlabelalt{EX label: 0}
        \end{tcolorbox}
    \end{minipage}
    \hfill 
    \begin{minipage}[t]{\textwidth}
        \begin{tcolorbox}[smallcasebox={Critique Response}]
            \textbf{<think>} \\
            1. Did I use the correct tables for the query? \\[1pt]
            - Yes, the superpower table contains the power\_name and id columns, which are relevant to the question about the power ID of cryokinesis. \\ [2pt]
            
            2. Did I correctly specify the column to select? \\[1pt]
            - Yes, the id column is the correct column to select as it represents the power ID. \\[2pt]
            
            3. Did I correctly filter the data based on the question? \\[1pt]
            - Yes, the WHERE clause correctly filters the rows where power\_name is 'cryokinesis', which is the power mentioned in the question. \\[2pt]
            
            4. Did I include any unnecessary columns or calculations? \\[1pt]
            - No, the query is straightforward and only selects the id column, which is what the question asks for. \\[2pt]
            
            5. Have I ensured that my query accurately targets the required data without adding unnecessary complexity? \\[1pt]
            - Yes, the query is simple and directly answers the question without any unnecessary complexity. \\[2pt]
            \textbf{</think>} \\
            
            \textbf{<result>} True \textbf{</result>} 
        \end{tcolorbox}
    \end{minipage}
    \caption{False negative example of EX scoring errors, along with their corresponding critique response generated by our \modelname~model.}
    \label{fig:case2}
\end{figure*}

\begin{figure*}
    \begin{minipage}[t]{\textwidth}
        \begin{tcolorbox}[casebox={Wrong Value but  Same Execution Results}]
            \rndlabel{Question} Among the players with an overall rating between 60 to 65, how many players whose going to be in all of your attack moves instead of defensing? \\
            
            \rndlabel{Pred SQL} 
            \begin{lstlisting}[style=sqlstyle, label=case3:pred_sql]
SELECT COUNT(*) 
FROM player_attributes 
WHERE overall_rating BETWEEN 60 AND 65 
AND attacking_work_rate != 'defensive' 
AND defensive_work_rate = 'low'    \end{lstlisting} 
            \rndlabel{Gold SQL} 
            \begin{lstlisting}[style=sqlstyle, label=case3:gt_sql]
SELECT COUNT(id) 
FROM 
Player_Attributes 
WHERE overall_rating BETWEEN 60 AND 65 
AND defensive_work_rate = 'low' \end{lstlisting}
            \rndlabel{Execution Result of Pred SQL}
            [(3594,)] \\[2pt]
            \rndlabel{Execution Result of Gold SQL}
            [(3594,)] \\[2pt]
            \rndlabelalt{EX label: 1}
        \end{tcolorbox}
    \end{minipage}
    \hfill 
    \begin{minipage}[t]{\textwidth}
        \begin{tcolorbox}[smallcasebox={Critique Response}]
            \textbf{<think>} \\
            1. Did I use the correct table for the query? \\[1pt]
  - Yes, the player\_attributes table contains the overall\_rating, attacking\_work\_rate, and defensive\_work\_rate columns, which are relevant to the question. \\[2pt]
    2. Did I correctly filter the players based on their overall rating? \\[1pt]
  - Yes, the condition overall\_rating BETWEEN 60 AND 65 correctly filters players with an overall rating between 60 and 65. \\[2pt]
    3. Did I correctly interpret the question regarding attacking and defensive work rates? \\[1pt]
  - The question asks for players who are ``going to be in all of your attack moves instead of defensing.'' This implies that the players should have a high attacking work rate and a low defensive work rate. \\[2pt]
  - The predicted SQL uses attacking\_work\_rate != `defensive' and defensive\_work\_rate = `low'. \textcolor{red}{However, the condition attacking\_work\_rate != `defensive' is not specific enough. It should be attacking\_work\_rate = `high' to ensure the player is focused on attacking.} \\[2pt]
4. Did I correctly count the players? \\[1pt]
  - Yes, the COUNT(*) function is correctly used to count the number of players that meet the specified conditions. \\[2pt]
5. Have I ensured that my query accurately targets the required data without adding unnecessary complexity? \\[1pt]
  - The query is straightforward and correctly targets the required data, but the condition for attacking\_work\_rate needs to be more specific.\\[2pt]
            \textbf{</think>} \\
            
            \textbf{<result>} \textcolor{red}{False} \textbf{</result>}  \\[2pt]

            \textbf{<correctedSQL>}
            \begin{lstlisting}[style=sqlstyle, label=case3:corrected_sql]
SELECT COUNT(*) 
FROM player_attributes 
WHERE overall_rating BETWEEN 60 AND 65 
AND attacking_work_rate = 'high' 
AND defensive_work_rate = 'low' \end{lstlisting}
            \textbf{</correctedSQL>}
        \end{tcolorbox}
    \end{minipage}
    \caption{False positive example of EX scoring errors, along with their corresponding critique response generated by our \modelname~model.}
    \label{fig:case3}
\end{figure*}

\newpage
\twocolumn
\section{Prompt Design}

\subsection{Prompt for LLM-based Process Reward Model}

\begin{lstlisting}[breaklines=true,label={prompt:process_judge},caption={Prompt for the rubric process judging.}]
Evaluate the correctness of a predicted SQL query and a provided correction SQL query for a given user query and schema by analyzing the reasoning process. For each step in the reasoning process, which includes a Question and an Answer, separately assess the correctness of the Question and the Answer. Additionally, verify whether the correction SQL query correctly satisfies the user query and conforms to the given schema. Provide justification for all assessments and summarize the step numbers of incorrect Questions and Answers, as well as the correctness status of the correction SQL, in a JSON format.

## Steps

1. **Understand the User Query and Schema**: Thoroughly understand the user query and the provided schema to form the basis for evaluating both the predicted and correction SQL queries.

2. **Analyze the Predicted SQL**: Review the predicted SQL query to ensure it aligns with the user query and schema.

3. **Analyze the Correction SQL**: Evaluate the correction SQL query to determine if it correctly answers the user query and is valid with respect to the schema. Check for syntactic correctness, semantic alignment with the user query, and schema compliance.

4. **Evaluate the Reasoning Process**: For each step in the reasoning process:
   - **Question**: Determine if the Question is relevant and correctly framed based on the user query and schema.
   - **Answer**: Assess if the Answer correctly addresses the Question and aligns with the predicted SQL.

5. **Provide Justification**: For each Question and Answer, provide a detailed justification for your assessment of its correctness. Also, justify your evaluation of the correction SQL query's correctness.

6. **Summarize Errors and Correction SQL Status**: Identify and list the step numbers of any incorrect Questions and Answers. Indicate whether the correction SQL query is correct or incorrect.

## Output Format

- Provide a JSON object summarizing the incorrect steps and the correction SQL evaluation:
  ```json
  {
    "incorrect_answers": [step_numbers],
    "incorrect_answer_ratio": 0.xx,    // number of incorrect answers / total reasoning steps, rounded to two decimals
    "correction_sql_correct": true|false,  // true if correction SQL correctly satisfies the user query and schema, else false
    "correction_sql_justification": "[detailed justification of the correction SQL correctness]"
  }
  ```

## Examples
**Example 1:**

- **User Query**: "Find the names of all employees in the 'Sales' department."
- **Schema**: Tables: Employees, Departments. Columns: Employees (id, name, department_id), Departments (id, name).
- **Predicted SQL**: `SELECT Employees.name FROM Employees JOIN Departments ON Employees.department_id = Departments.id WHERE Departments.name = 'Sales';`
- **Correction SQL**: `SELECT name FROM Employees WHERE department_id IN (SELECT id FROM Departments WHERE name = 'Sales');`
- **Reasoning Process**:
  - **Step 1**:
    - **Question**: "What tables are involved in the query?"
    - **Answer**: "Employees and Departments."
  - **Step 2**:
    - **Question**: "How are the tables related?"
    - **Answer**: "Through the department_id in Employees and id in Departments."
  - **Step 3**:
    - **Question**: "What condition filters the results?"
    - **Answer**: "Departments.name = 'Sales'."

- **Assessment**:
  - **Incorrect Answers**: []
  - **Correction SQL Correctness**: true
  - **Justification**: The correction SQL correctly retrieves employee names from the Employees table where the department matches 'Sales' by using a subquery on Departments, which aligns with the user query and schema.

- **Output**:
  ```json
  {
    "incorrect_answers": [],
    "incorrect_answer_ratio": 0.00,
    "correction_sql_correct": true,
    "correction_sql_justification": "The correction SQL correctly retrieves employee names from Employees where department_id matches Departments with name 'Sales', aligning with the user query and schema."
  }
  ```

## Notes

- Ensure the reasoning process is logical and follows a clear sequence.
- Pay close attention to the alignment between the user query, schema, predicted SQL, and correction SQL.
- Consider edge cases such as ambiguous column names, complex joins, or nested queries.
- The correction SQL correctness check should include syntactic validity, semantic correctness, and schema compliance.
- Output only the JSON result as specified, without additional commentary.
\end{lstlisting}

\subsection{Prompt Used for Critique Model Training and Inference}
\label{sec:prompt_inference}

\begin{lstlisting}[breaklines=true,label={prompt:critique},caption={Prompt for the training and inference of critique model.}]
## ROLE
    You are an expert in identifying code errors in the NL2SQL (Natural Language to SQL) task. Based on the provided question, table structure (schema), and other information, determine whether the corresponding code (Predicted SQL) is correct.  
    Please follow the steps below:
    - self ask step by step reasoning
    - Determine whether the SQL solution is correct based on the reasoning results
    - If there are errors, please correct the mistakes in the SQL code and provide the correct SQL code

## Example of Asking myself: step-by-step reasoning

    INPUT:
    - Question: How many students, on average, does each college have enrolled?

    - Predicted SQL: 
    ```sql
    select avg(college.enr), count(*) from college group by college.enr
    ```

    - Schema Overview:
    Table schema: There are 1 tables:college.
    the structure of table college is as follows:
    | column name  |  column type |
    | ------------ | ------------ |
    | """"cName"""" | text |
    | state | text |
    | enr | number |
    """"cName"""" is the primary key.


    OUTPUT:
    ```
    <think>
    1. **Did I use the correct table for the query?**
        - Yes, the `College` table contains the `enr` column which represents the enrollment number, so it is the correct table to use for calculating the average enrollment.
    2. **Did I correctly specify the column to calculate the average?**
        - Yes, using `avg(college.enr)` is appropriate as we want to find the average number of students enrolled in each college, and `enr` is the column that stores the enrollment data.
    3. **Did I include any unnecessary columns or calculations?**
        - The predicted SQL includes `count(*)`, which is not relevant to the question of finding the average enrollment per college. The question only asks for the average enrollment, so this `count(*)` should be removed.
    4. **Did I use the correct grouping?**
        - The `group by college.enr` in the predicted SQL is incorrect. Grouping by the enrollment value doesn't make sense for this problem. We just need to calculate the average of all the enrollment values in the `College` table, so no grouping is required.
    5. **Have I ensured that my query accurately targets the required data without adding unnecessary complexity?**
        - The original query adds unnecessary complexity with the `count(*)` and the incorrect `group by` clause. The query should be simplified to just calculate the average enrollment.
    </think>
    <result>False</result>
    <correctionSQL>
    ```sql
    SELECT AVG(college.enr) FROM college
    ```
    </correctionSQL>
    ```


## Output Format:
    ```
        <think> Your step-by-step process of thinking and reasoning  </think>
        <result>If mistakes are found during your thinking and reasoning process, give False; otherwise, give True </result>
        <correctionSQL> A valid SQL which is either corrected SQL or the predicted SQL without any change if you think it is correct already </correctionSQL>
    ```
\end{lstlisting}

\end{document}